\documentclass[letterpaper, 10 pt, conference]{ieeeconf}  

\IEEEoverridecommandlockouts                              

\usepackage{graphics} 
\usepackage{epsfig} 
\usepackage{mathptmx} 
\usepackage{times} 
\usepackage{amsmath} 
\usepackage{amssymb}  
\usepackage{hyperref} 
\usepackage{xcolor} 
\usepackage{arydshln} 
\usepackage{color} 
\usepackage{float}

\definecolor{somegray}{rgb}{0.5, 0.5, 0.5}
\newcommand{\darkgrayed}[1]{\textcolor{somegray}{#1}}
\makeatletter
\newcommand*\titleheader[1]{\gdef\@titleheader{#1}}
\AtBeginDocument{%
  \let\st@red@title\@title
  \def\@title{%
    \vskip-3em
    \bgroup\normalfont\large\centering\@titleheader\par\egroup
    \vskip1.5em\st@red@title}
}
\makeatother

\titleheader{\darkgrayed{
This paper has been accepted for publication at the IEEE/RSJ International Conference on Intelligent
Robots and Systems (IROS), Prague, 2021.
\copyright IEEE}}

\title{\LARGE \bf
Powerline Tracking with Event Cameras
}

\author{Alexander Dietsche, Giovanni Cioffi, Javier Hidalgo-Carri\'o, Davide Scaramuzza
\thanks{The authors are with the Robotics and Perception Group, Department of Informatics, University of Zurich, and Department of Neuroinformatics, University of Zurich and ETH Zurich, Switzerland (\protect\url{http://rpg.ifi.uzh.ch}).
This work was supported by the National Centre of Competence in Research (NCCR) Robotics through the Swiss National Science Foundation (SNSF), and by the European Union’s Horizon 2020 Research and Innovation Programme under grant agreement No. 871479 (AERIAL-CORE).}
}

\begin{document}

\maketitle
\thispagestyle{empty}
\pagestyle{empty}

\begin{abstract}
Autonomous inspection of powerlines with quadrotors is challenging. Flights require persistent perception to keep a close look at the lines. We propose a method that uses event cameras to robustly track powerlines. Event cameras are inherently robust to motion blur, have low latency, and high dynamic range. Such properties are advantageous for autonomous inspection of powerlines with drones, where fast motions and challenging illumination conditions are ordinary. 
Our method identifies lines in the stream of events by detecting planes in the spatio-temporal signal, and tracks them through time. The implementation runs onboard and is capable of detecting multiple distinct lines in real time with rates of up to $320$ thousand events per second. The performance is evaluated in real-world flights along a powerline. The tracker is able to persistently track the powerlines, with a mean lifetime of the line $10\times$ longer than existing approaches.
\end{abstract}

\vspace{0.3cm}


\section*{Supplementary material}
\label{sec:supplementary_material}

A video showing an application of our powerline tracker is available at~\url{https://youtu.be/KnBJqed5qDI}. The code can be found here~\url{https://github.com/uzh-rpg/line_tracking_with_event_cameras}.

\section{Introduction}
\label{sec:intro}

Quadrotors are a fast-to-deploy and a cost-effective solution for powerline inspection~\cite{DroneiiWP18}. To perform such operation, teams of specialized human operators typically use ropes or scaffoldings to access the powerline infrastructure and manned helicopters for long range inspection. This task requires months of training, is time consuming, and entails high risk for humans. Quadrotors can cut costs by $50\%$~\cite{DroneiiWP18}, reducing the in-person operation, and improving the data collection. Also, they can drastically reduce the inspection time and increase the frequency and quality of inspection.

Current solutions are transferring from research to industry by partially helping humans in the most challenging scenarios~\footnote{\url{https://aerial-core.eu}}~\cite{DroneiiWP18, DanishUAS2016}.
However, they require a direct visual line of sight and that the quadrotor follows a predefined path to inspect a dedicated zone. The energy sector, in general, and the powerline sector, in particular, raise the fundamental need for closed-loop solutions with highly coupled perception and action~\cite{falanga2018pampc}. 

\begin{figure}[H]
    \centering
    \includegraphics[width=1.0\linewidth] {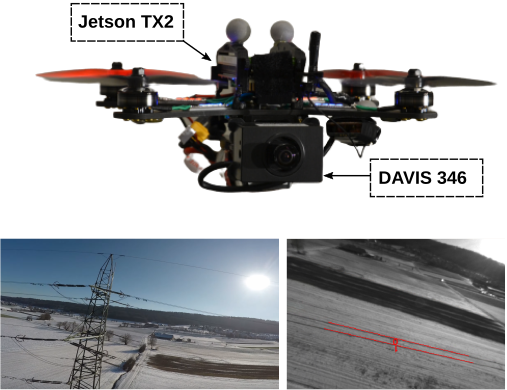}
    \caption{Quadrotor with the onboard computer and the event camera (top figure). Bottom figures show the real test scenario during the outdoor experiments. Bottom left: color image with a power line mast. Bottom right: two lines tracked and marked in red with their unique ID number.}
    \label{fig:eyecatcher}
\end{figure}

Robust and persistent line tracking is beneficial for perception-aware planning~\cite{costante2016arxiv, zhang2018icra}. The perception-aware planner uses the information on the position of the line to compute a path that guides the drone to its target location while keeping the desired powerline in the camera field of view. This allows recording images of the powerlines that can be analyzed offline to evaluate the health of the powerline infrastructure.

Onboard perception for quadrotors is currently limited by the payload, where a significant portion of it is used for high-resolution sensors, such as LiDAR and high definition cameras.
These sensors are mainly required for data acquisition, high level planning, and offline post-processing, increasing latency when in use by the low level controller. A quadrotor needs low-latency sensory information to perform closed-loop navigation and guarantee safety and quality during inspection.

Event cameras have gained tremendous attention in the last decade in robotics, as well as in computational photography and computer vision~\cite{Gallego20pami}. Their inherent robustness against motion blur, low latency, and high dynamic range~\cite{sun2021ral, falanga2020scirob} is attractive for robotic applications. These properties are beneficial for powerline inspection tasks in case the drone needs to perform agile manoeuvres, for example to avoid powerline masts.
However, due to their unconventional output, event cameras also present new challenges, and tracking of objects, such as lines, is a difficult task due to the sparsity and frequent changes in appearance in the stream of the events.

The difficulty of line tracking using events led previous works~\cite{dimitrova2020towards, eguiluzasynchronous, tschopp2021hough2map, von2012lsd, brandli2016elised, gentil2020idol, everding2018low} to explore different techniques but pursuing a common goal: robustness. We identify the robustness of tracking a line as the combination of three desired characteristics. (1) \textit{correctness} (i.e., the capability of solely tracking lines), (2) \textit{instance} (i.e., the capability of tracking lines with a unique identifier, and (3) \textit{persistence} (i.e., how long the line can be tracked). These characteristics are summarized in Table~\ref{tab:comparison}.

In this work, we present a method that captures the aforementioned characteristics to robustly track lines for powerline inspection.\\
The main contributions of this work are:
\begin{itemize}
    \item a \textit{spatio-temporal} event-based line tracker optimized for powerline inspection. Our algorithm extends~\cite{everding2018low} with novel processing blocks. 
    \item the use of \textit{hibernation} to improve the persistence of the line tracker by a factor of 10 compared to~\cite{everding2018low}.
    \item a real-time implementation capable to run onboard a lightweight resource-constrained quadrotor platform.
    \item experiments on a real world powerline dataset showing superior performance than a custom version of~\cite{everding2018low}.
    \item the release of the code fully open source.
\end{itemize}

\begin{table}[t]
\centering

\begin{tabular}{|l|c|c|c|}
\hline
\textbf{Method category} & \multicolumn{1}{l|}{\textbf{Correctness}} & \textbf{Instance}        & \multicolumn{1}{l|}{\textbf{Persistence}} \\ \hline
Hough transf.~\cite{dimitrova2020towards, eguiluzasynchronous}, \cite{tschopp2021hough2map} & \textcolor{green}{+}  &  \textcolor{red}{-}  &  \textcolor{green}{+}    \\ \hline
Non parametric~\cite{von2012lsd, brandli2016elised} &  \textcolor{green}{+}  & \textcolor{red}{-}  &  \textcolor{red}{-}  \\ \hline
Spatio-temporal~\cite{gentil2020idol, everding2018low} &  \textcolor{green}{+}  & \textcolor{green}{+}  &  \textcolor{red}{-}                   \\ \cdashline{1-4}
Ours            & \textcolor{green}{+}  &  \textcolor{green}{+}  &  \textcolor{green}{+}                   \\ \hline
\end{tabular}

\caption{Existing methods and their categories. Our approach belongs to the spatio-temporal category adding persistence with hibernation.}
\label{tab:comparison}
\vspace{-0.5cm}
\end{table}
\section{Related Work}
\label{sec:soa}

The recent literature on event-based line trackers can be divided into three categories: 

\textit{Hough transform based methods}: 
These methods use polar coordinate to parameterize the lines~\cite{duda1972use} and search for the lines that contain the largest number of events in resulting parameter space. 
A Hough transform based line tracker was used for low-latency and high-bandwidth control of a one dimensional dualcopter in~\cite{dimitrova2020towards}.
The tracker ran on a sliding window of events, as proposed in~\cite{mueggler2014event}, and was used in combination with a Kalman filter to estimate the dualcopter state.
The work in~\cite{eguiluzasynchronous} proposes an application of a Hough transform based tracker for agile aerial robot maneuvering.
Consecutive Hough transform based detectors were used in~\cite{tschopp2021hough2map} to detect and track poles in a railway localization system.\\
Although Hough transform based methods have been used for events, they do not fully leverage the asynchronous nature of event cameras. 
Performing the projection from the pixel to the parameter space and searching such parameter space on an event-by-event basis is computationally costly and requires fine adjustment of the peak selection and binning of the parameters.
Using a sliding window of events helps in reducing the computational cost but introduces latency. 

\textit{Non parametric based methods}:
These methods includes a line detector which calculates the spatial derivative of every pixel and clusters the pixels into support regions based on their orientation.
The support regions are then used to calculate the line segments~\cite{von2012lsd}.
This approach minimizes the need of parameter tuning.
A non parametric method was adopted by the event-based line segment detector ELiSeD~\cite{brandli2016elised}.
ELiSeD attributes an orientation to an incoming event and then clusters together those events that have similar orientation. 
This method has the tendency to produce small line segments that are short-lived. 
In our task, the lines of interest have large length and are mostly long-lived.

\textit{Spatio-temporal based methods}:
These methods make the assumption that lines travel approximately at a fixed speed in small time intervals. When such assumption holds, lines describe a plane in the spatio-temporal event space. 
In~\cite{gentil2020idol}, the normal vectors in the spatio-temporal space were computed for each incoming event using a local neighborhood.
Events with similar normal vectors were clustered together to form lines.
The detected lines and inertial measurements were used to estimate the camera poses in an event-based visual odometry algorithm.
A spatio-temporal based line tracker algorithm was also proposed in~\cite{everding2018low}. 
The algorithm was able to track slowly translating lines for long time.
Moreover, each detected line had an internal state and could be uniquely identified.
Since these characteristics are desirable for powerline tracking applications, we built our line tracker upon~\cite{everding2018low}.
\section{Methodology}
\label{sec:method}

\begin{figure}[t]
    \centering
    \includegraphics[width=0.99\linewidth]{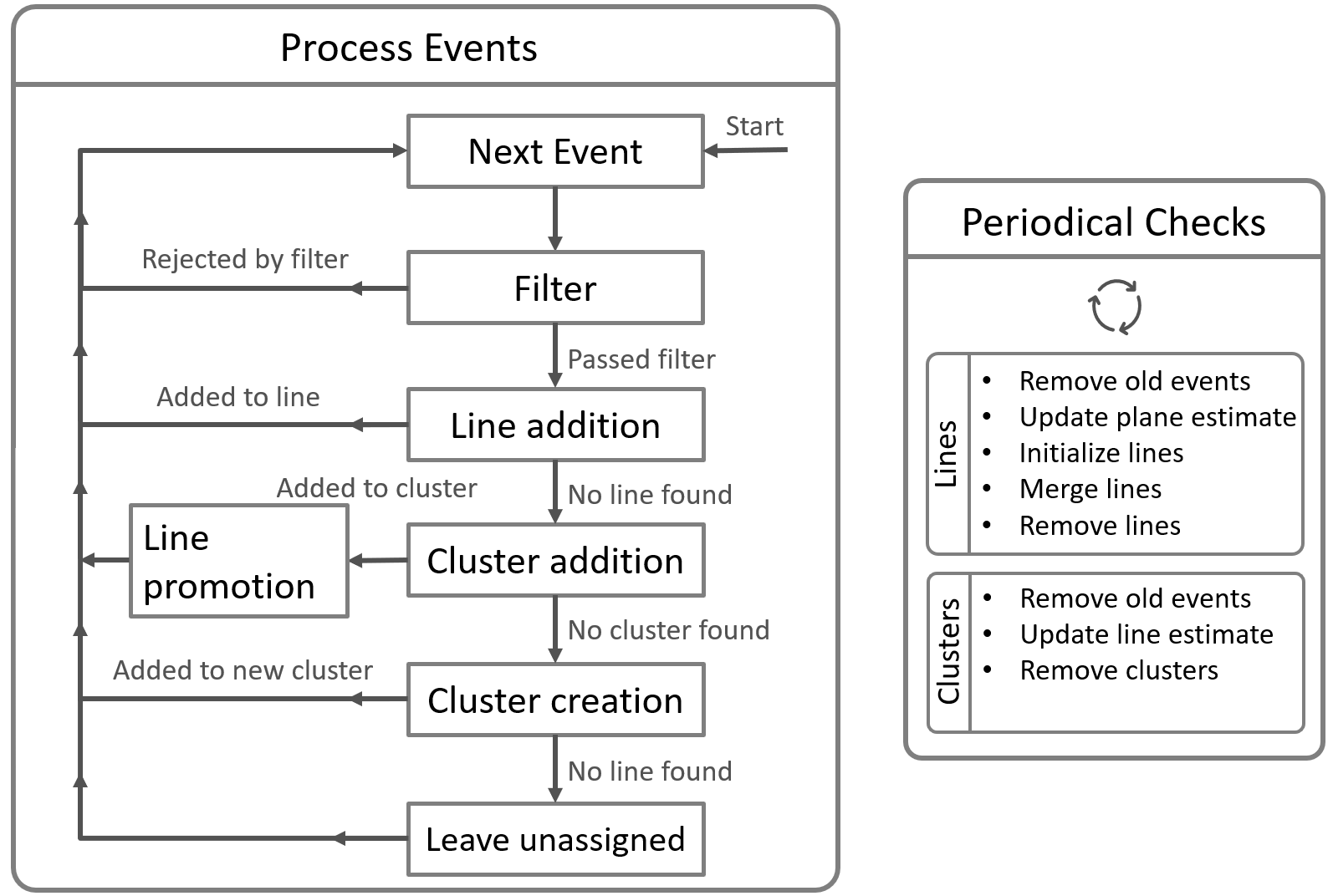}
    \caption{Overview of the proposed approach, which extends~\cite{everding2018low}.}
    \label{fig:approach}
\end{figure}

Our approach, based on~\cite{everding2018low}, fits planes in short time periods through the spatio-temporal space of the events. 
A line is detected if the plane fitting satisfies some requirements (see Sec.~\ref{sec:line_promotion}). 
We rely on the assumption that the velocity of the lines is approximately constant in the short time period. 
Such assumption holds for the lines whose motion mainly consists of translation.\\
The line tracking algorithm has four main steps. 
Events are filtered based on a refractory-time filter and a neighborhood filter. 
Each event that passes the filter becomes a candidate for being added to an existing line.
If the line addition fails, the event is a candidate for being added to a cluster. 
If there are not clusters available for such event, it is left unassigned.
New clusters can be generated from the unassigned events using the newest event as the starting point. 
An overview of the proposed algorithm can be seen in Fig.~\ref{fig:approach}. 
In parallel to the processing of new incoming events, the existing lines and clusters are periodically checked in a separate thread. 
These checks are necessary to periodically remove old events, lines, clusters, and update the line and cluster estimates.

\subsection{Filtering}\label{sec:filtering}

The initial event filtering consists of two independent steps.  
In the first step, all event firings at a given location are suppressed for a refractory period after an event has been fired at that position.
The refractory period is either 8 ms, if an event of the same polarity is fired, or 1 ms, if an event of different polarity is fired.
In this way, the filter suppresses spurious event firings of the same polarity.  
In the second step, the filter checks the $5\times5$ pixel by 70 ms volume surrounding the incoming event.
If there are at least three other events registered in that volume, the event passes the filter.   
Both filters are implemented using the Surface of Active Events (SAE)~\cite{Mueggler17BMVC}. 
The SAE is comparable with a grayscale image, where instead of intensity values, the newest timestamps are stored. 
In the refractory time-based filter, we get the newest timestamp from the SAE and compare it with the timestamp of the incoming event. 
In the neighborhood filter, we get the timestamp of the neighboring events and count the number of timestamps not older than 70 ms.

\subsection{Line Addition}\label{sec:line_addition}

\begin{figure}[tb]
    \centering
    \includegraphics[width=0.6\linewidth]{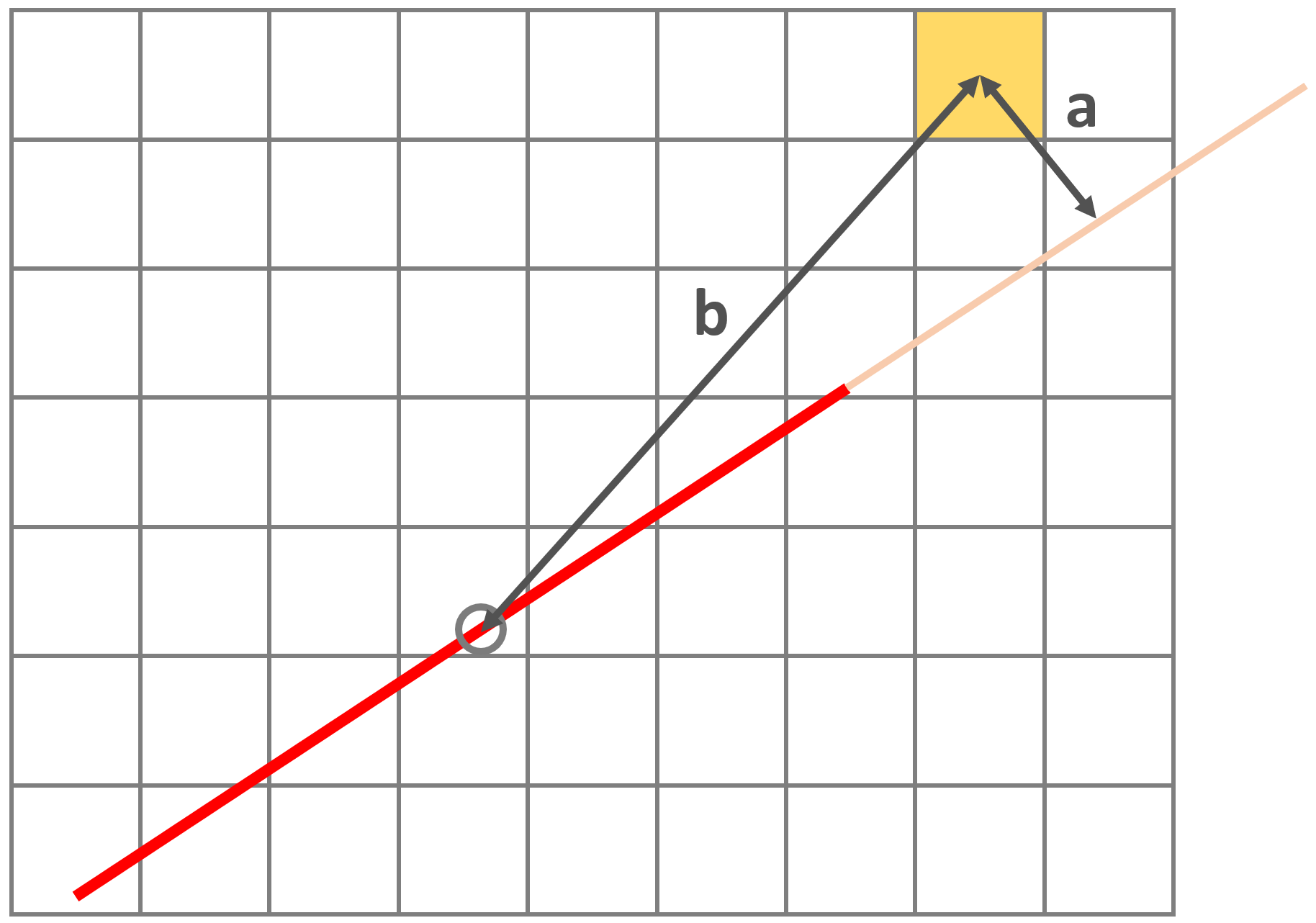}
    \caption{Distances $a$ and $b$ are calculated to check if the incoming event (the yellow square) can be added to the line.}
    \label{fig:line_addition}
\end{figure}

Events that pass the initial filtering process can be added to the existing lines. 
We evaluate if an incoming event belongs to a line by computing its distances to the line, $a$, and its distance to the center of the line, $b$, as depicted in Fig.~\ref{fig:line_addition}.
The event is added to the line if the distance $a$ is below 1.8 pixel, and the distance $b$ is below the line length. 
If multiple lines fulfill this condition, the event is ignored.

\subsection{Cluster Addition}\label{sec:cluster_addition}

The cluster addition step tries to add the incoming event, which has not been assigned to a line, to an existing cluster. 
This step is similar to the line addition step and depends on the same two distances. 
The distance to the midpoint and the distance to the inferred line must be below a certain threshold for the event to be added to the cluster.
If multiple clusters fulfill these conditions, we check if they can be merged. 
This check consists of calculating the relative angle of the respective inferred lines.  
If such angle lies below 15 deg, the clusters are merged.
If the event is successfully added to a cluster and that particular cluster has collected more than 35 events, the cluster can be promoted to a line.  

\subsection{Line Promotion}\label{sec:line_promotion}

\begin{figure}[tb]
    \centering
    \includegraphics[width=0.7\linewidth]{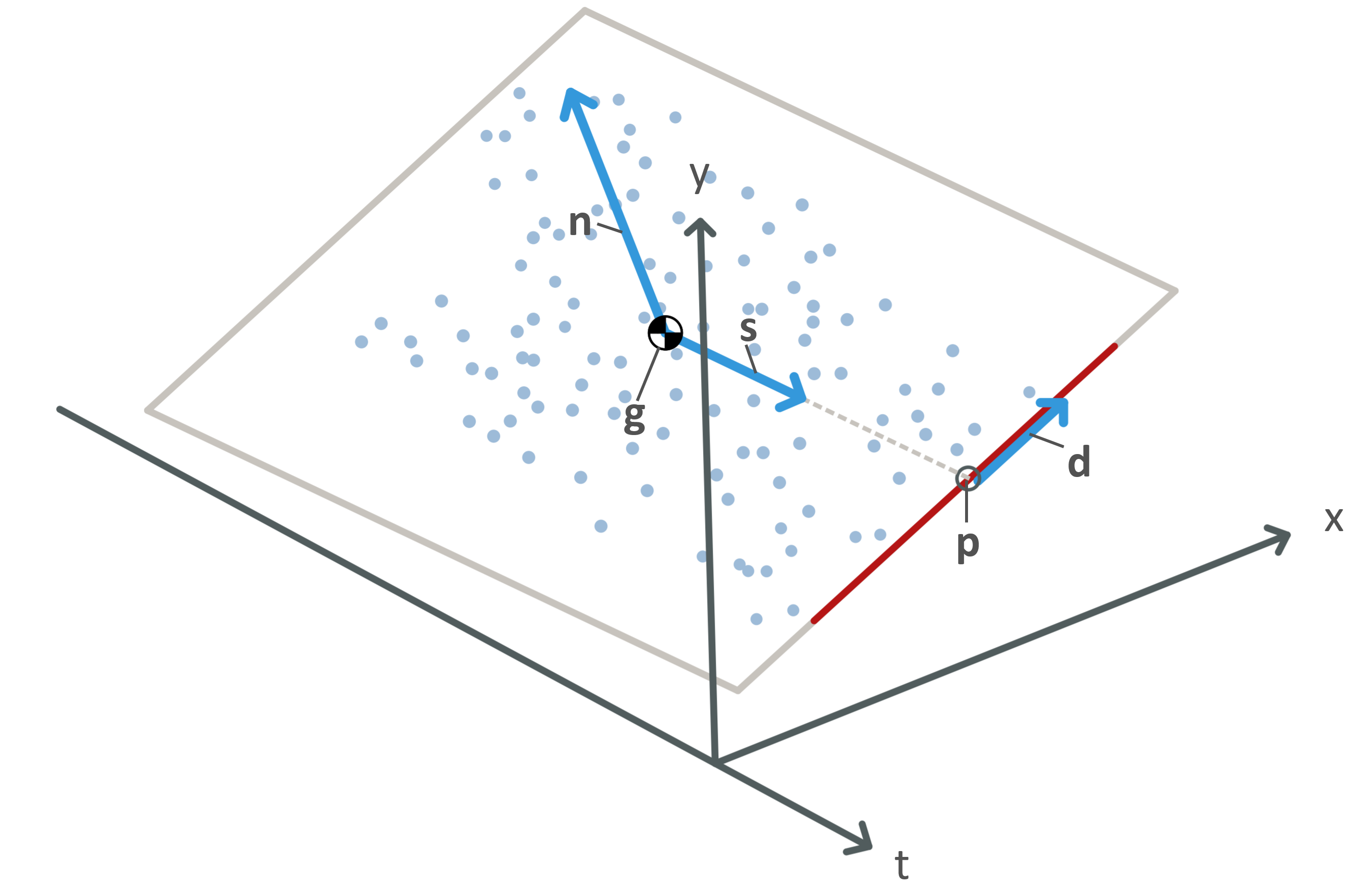}
    \caption{
    The detected line, in red, is the intersection of the fitted plane and the $xy$ plane at the current time. 
    The point $\mathbf{p}$ denotes the midpoint of the line, which is the projection of the center of gravity of the events, $\mathbf{g}$, along the direction $\mathbf{s}$ onto the line. The vector $\mathbf{n}$ is the plane normal, and the vector $\mathbf{d}$ is the direction of the line.
    }
    \label{fig:plane_fitting}
\end{figure}

A cluster is promoted to a line if the events belonging to that cluster lie on a plane.
The covariance matrix of the event coordinates is computed, and if its smallest eigenvalue is below a certain threshold, the cluster is promoted to a line. 
To calculate the covariance matrix, the coordinates are centered to have zero mean.
Let $\mathbf{X} \in \textbf{R}^{\text{N} \times 3}$ be the matrix that contains the centered coordinates of the $\text{N}$ events that belong to the cluster.
We then perform the eigen-decomposition of the covariance matrix $\mathbf{C} = \frac{1}{\text{N}-1}\mathbf{X}^\intercal\mathbf{X}=\mathbf{Q}\Lambda\mathbf{Q}^\intercal$.
Since the covariance matrix is symmetric and positive semi-definite, $\Lambda$ is guaranteed to contain positive real eigenvalues. 
If the smallest eigenvalue is smaller than 1.2 pixel the cluster will be promoted to a line.  
The fitted plane is located at the center of gravity of the events, with its normal $\mathbf{n}$ being the eigenvector corresponding to the smallest eigenvalue. 
The detected line is then obtained by calculating the intersection of the fitted plane with the $xy$ spatial plane at current time $t$. 
We are interested in the line direction $\mathbf{d}$, the midpoint $\mathbf{p}$, and the line length $l$. 
The direction $\mathbf{d}$ is the intersection of the $xy$ plane at $t$ and the fitted plane. 
The $xy$ plane has normal $\mathbf{e}_t = [0 \;\; 0 \;\; 1]$ and the fitted plane has normal $\mathbf{n}= [n_x \;\; n_y \;\; n_t]$. 
The line direction is then calculated as:
\begin{equation}\label{eq:line_distance}
\mathbf{d} = \mathbf{n} \times \mathbf{e}_t = 
\begin{bmatrix} 
n_2 & -n_1 & 0 
\end{bmatrix}^\intercal.
\end{equation}
To calculate the midpoint $\mathbf{p}$, we project the center of gravity $\mathbf{g}$ of the events onto the line. 
The vector $\mathbf{s}$ which points from $\mathbf{g}$ to $\mathbf{p}$ lies within the fitted plane and is thus perpendicular to the normal vector $\mathbf{n}$. 
The vector $\mathbf{s}$ is also perpendicular to $\mathbf{d}$, since the projection of a point onto a line is always orthogonal to the line direction, and it is computed as:
\begin{equation}\label{eq:line_midpoint1}
\mathbf{s} = \mathbf{n} \times \mathbf{d} = 
\begin{bmatrix}
n_x n_t & n_y n_t & -n_x^2 -n_y^2
\end{bmatrix}^\intercal. 
\end{equation}
The point $\mathbf{p}$ is then found as:
\begin{equation}\label{eq:line_midpoint2}
\mathbf{p} = \mathbf{g} + \frac{t - \overline{t}}{-n_x^2 - n_y^2} \mathbf{s},
\end{equation}
where $\frac{t - \overline{t}}{-n_x^2 -n_y^2}$ is a scaling factor.\\
The calculation of the line length is based on the assumption that the events are uniformly distributed along the line.
Using the formulation of the standard deviation of the Uniform distribution, $\Gamma$, $\sigma = \sqrt{\mathbb{E} \bigr[  \, \Gamma^2 \bigl] - ( \mathbb{E} \bigr[ \Gamma \bigl] )^2}$ and the fact that the data is centered (i.e., $\mathbb{E} \bigr[ \Gamma \bigl] = 0$), the line length is $l = \sqrt{12 \sigma}$.
We compute $\sigma$ by projecting the largest and second largest eigenvectors scaled by their eigenvalues onto the line.
Fig.~\ref{fig:plane_fitting} depicts the quantities involved in the line promotion step.

\subsection{Cluster Creation}\label{sec:cluster_creation}

\begin{figure}[tb]
    \centering
    \includegraphics[width=0.90\linewidth]{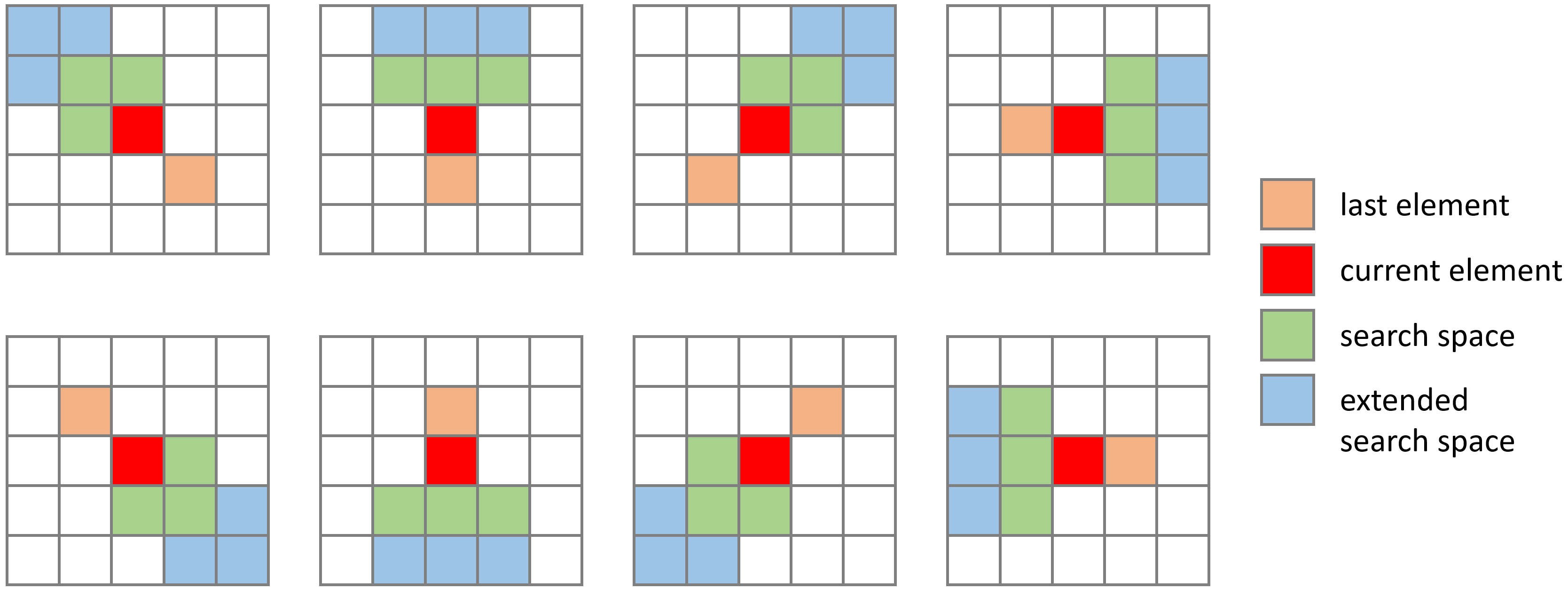}
    \caption{Directed chain growth depends on the relative position of the last and current chain element. If no event is found in the search space (green) the extended search space (blue) will be checked.}
    \label{fig:chain_growth}
\end{figure}

Events that are added neither to existing lines nor to clusters are left unassigned and are stored in a SAE. 
We use the event position as a starting point to find a chain of connected events in the SAE. There are two separate SAEs, one for each polarity.
The first chain element is the incoming event. 
The second chain element is the youngest event in the $3\times3$ neighborhood that is not older than a certain threshold. 
From there on, the chain grows in a directed fashion. 
Depending on the relative position of the last and current element, the next element is searched in a predefined search space. 
If no element is found, this space is extended.
The predefined search spaces are depicted in Fig.~\ref{fig:chain_growth}, where both the initial search pattern (green) and the extended search pattern (blue) are shown.
The chain growth is stopped once no more events are found or if the chain reaches a predefined length.

\subsection{Initialization}\label{sec:initialization}

We propose to add an initialization step in order to be more conservative about the line detection.
The proposed tracker does not make any distinction between lines that have many evenly distributed events and lines that have just few events that are far apart one another. 
In view of the powerline inspection task, we are more interested in the former ones.
To that end, newly created lines start out in the \textit{initializing} state and collect events during an initialization period.
We check the connected length at the end of the initialization period.
The connected length is calculated by projecting the events along the fitted plane onto the line.
The projected events are then binned into bins of size 2~pixel.
The connected length is the length of the longest non-empty bin chain with gaps of max one empty bin.
If the connected length is above 70~pixel, the line receives an unique line ID and goes into the \textit{active} state. 
Otherwise, the line is discarded. 
\begin{figure}[t!]
   \centering
   \includegraphics[width=0.8\linewidth]{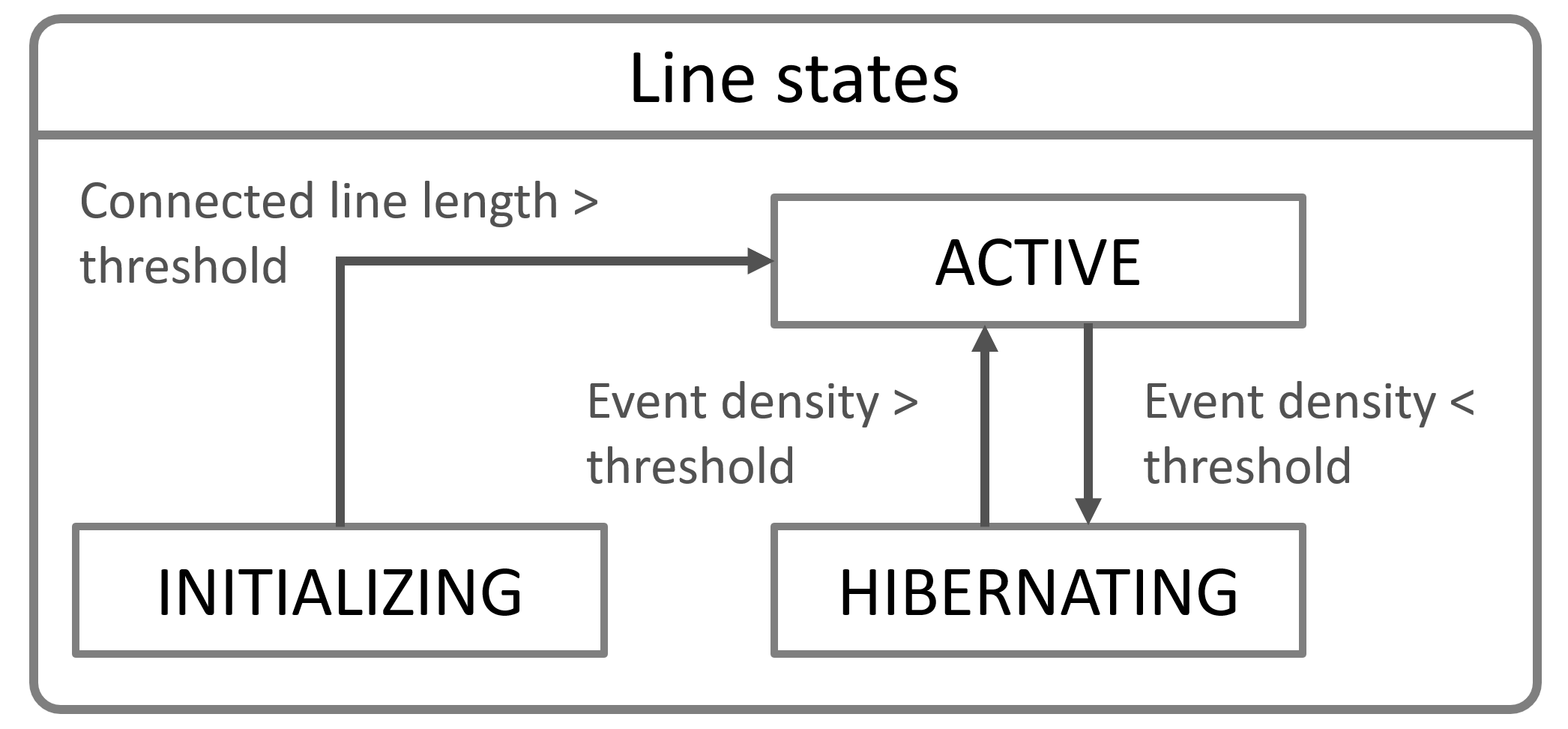}
   \caption{Line states and their transition conditions.}
   \label{fig:line_states}
\end{figure}

\subsection{Hibernation}\label{sec:hibernation}

Hibernation is introduced to make the line tracker robust against lines that change direction, which corresponds to change in appearance in the stream of events.
In the case of powerline tracking onboard a quadrotor, the change of direction of the lines is due to the flying path of the robot. 
At the moment of a change of direction, just few events are fired and
the last estimate of the event plane is not perfectly perpendicular to the $xy$ plane since events only up to a certain point in the past are considered. 
Consequently, for a short period of time, the estimated line keeps moving in the original direction. 
Once new events are fired, the estimated line has moved too far in the opposite direction and is lost. 
A new line instead will be detected.
Hibernation alleviates this problem by calculating the midpoint of the line as the orthogonal projection of the center of gravity of the events onto the current $xy$ plane instead of the projection onto the fitted plane.
Thus, a hibernated line stand still until it enters the \textit{active} state again, once new events are added.\\
Another issue that the tracker would face without hibernation is that in the case of slow direction changes no events are fired for an extended period.
The lines would lose many events since old events are removed periodically and only very few ones are added. 
Once a line contains just few events, it will be deleted. 
Hibernation helps in this case since old events of hibernated lines are not removed.\\
The condition for the line state to change between the \textit{active} and \textit{hibernated} state depends on the event density. 
The event density is the number of events younger than 25 ms divided by the length of the line.
If the event density is below 0.08 [$\frac{\text{\# events}}{\text{pixel}^2 \cdot \text{ms}}$], the line state changes from \textit{active} to \textit{hibernated}.

\subsection{Periodical Checks}\label{sec:periodical_checks}

The tracker performs periodical checks on the current clusters and lines.  
Events belonging to either clusters or lines that are older than 50 ms are removed.
Clusters whose the newest event is older than 40 ms are also deleted. 
For the remaining clusters, their inferred lines are updated by incorporating the newly added events. 
The periodical checks on the lines depend on the line state. 
For \textit{initializing} lines, we check if their initialization period has ended, in that case we compute their connected length. 
If the connected length exceeds 70 pixel, the line will change from state \textit{initializing} to \textit{active}. 

The periodical checks for \textit{active} lines consist of updating the fitted plane estimate and checking the line length.  If the line is too short, it is discarded. 
Finally, \textit{hibernated} lines are checked for the time they have been hibernated. 
If such time exceeds 1 second, the line is deleted.  
We also check for the age of the newest event, which has to be below 40 ms.  
Finally, we check if two lines can be merged.  
The merge conditions are fulfilled when the angle difference of the inferred lines is below 8.0 deg and the midpoint distances are below 3.5 pixel.  
If that is the case, the line with the lower ID will receive the events of the other line, which is deleted.

\subsection{Limitations}\label{sec:limitations}

The tracker performs at its best in scenes where the lines of  interest move in a mostly translating fashion. 
Fast rotating lines lead to detection problems due to the  underlying assumption that the lines form planes in the  spatio-temporal space.
In texture-rich environments where there are a lot of events fired but no obvious lines are present, the tracker has a tendency to detect spurious lines. 
Under these circumstances, there are enough events for lines to pass initialization step and to continue existing.  
With a steady flow of incoming events, the tracker see no reason to remove the lines. Hibernation exacerbates this issue by preventing the tracker from quickly deleting these lines.
However, this is not an usual case during powerline inspection tasks.\\
Our tracker may not be able to track the line when only few events are fired. 
This occurs in the case when the drone flies perfectly parallel to the powerline.
\section{Experiments}
\label{sec:results}

\subsection{Powerline dataset}\label{sec:datasets}
We evaluated the proposed line tracker on a dataset recorded onboard a manually piloted quadrotor tracking powerlines. 
The quadrotor was equipped with 
an iniVation Dynamic and Active Vision Sensor (DAVIS) 346 event camera.
The camera resolution was $346 \times 260$ pixel.
A Matrix Vision MLC200wG Bluefox camera was also mounted to record higher quality standard frames than those provided by the DAVIS camera.
The frames of the Bluefox camera were only used for visualization and evaluation purposes. 
The main parameters used for the tracker are in the Appendix (see Sec.~\ref{sec:appendix}).

\subsubsection{Evaluation} \label{sec:qualitative_evaluation}

\begin{figure}[tb]
    \centering
    \includegraphics[width=0.493\linewidth]{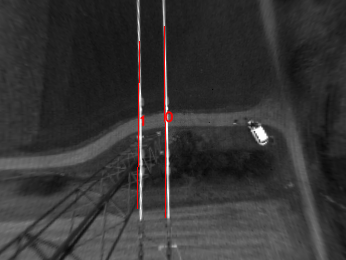} \hfill
    \includegraphics[width=0.493\linewidth]{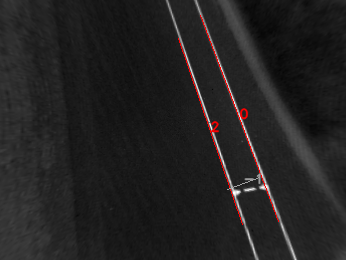} 
    \\[\smallskipamount]
    \includegraphics[width=0.493\linewidth]{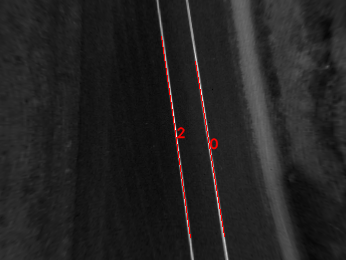} \hfill
    \includegraphics[width=0.493\linewidth]{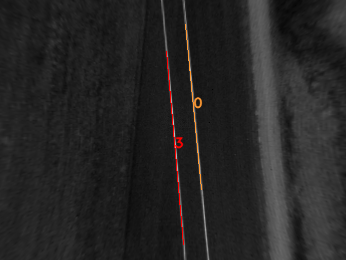}
    \caption{Detected lines on the powerline dataset recorded onboard a quadrotor. 
    Red lines are in the \textit{active} state, yellow lines are in the \textit{hibernated} state, and grey lines are in the \textit{initializing} state.}
    \label{fig:power_line_with_hib}
\end{figure}

Since, to the best of our knowledge, it does not exist an event-based dataset with ground truth for powerlines, we used the frames captured by the Bluefox camera to qualitatively evaluate the accuracy of the proposed line tracker. 
This was done by overlaying the detected lines on the grayscale standard images, see Fig.~\ref{fig:power_line_with_hib}. 
The tracker was able to successfully extract lines from the event stream and keep track of them. 
The detected lines are shown at different times during the flight in Fig.~\ref{fig:power_line_with_hib}. 
The tracker shows good persistence, i.e., once a line is detected, it is tracked for a long period of time. 
This result can be visualized in the accompanying video by the mostly constant line IDs throughout the sequence. 
The few cases where the lines are lost are due to very few event firings which occur when the drone flies almost parallel to the line.
This problem could be addressed by combining the event-based with a standard frame-based line tracker, using an approach similarly to the one proposed in~\cite{vidal2018ultimate}.
The tracker persistence is quantified by the mean lifetime of a line. 
For the flight in Fig.~\ref{fig:power_line_with_hib}, the mean lifetime was 31.12 s.
\subsubsection{Benefits of Hibernation}\label{sec:hibernation_polarity}

\begin{figure}[tb]
    \centering
    \includegraphics[width=0.493\linewidth]{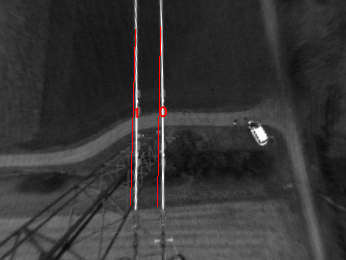} \hfill
    \includegraphics[width=0.493\linewidth]{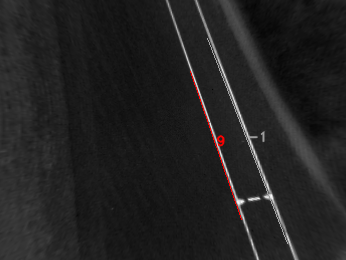} 
    \\[\smallskipamount]    
    \includegraphics[width=0.493\linewidth]{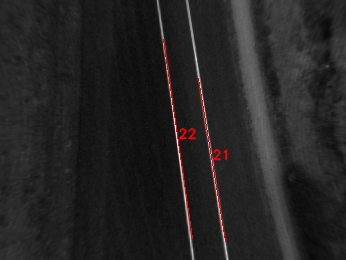} \hfill
    \includegraphics[width=0.493\linewidth]{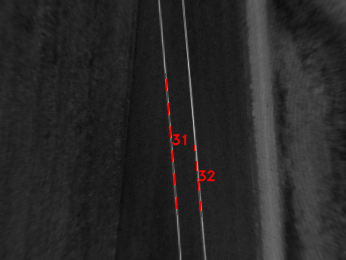} 
    \caption{Detected lines on the powerline dataset without hibernation. The images are the same as in Fig. \ref{fig:power_line_with_hib}.}
    \label{fig:power_line_no_hib}
\end{figure}
To quantify the impact of hibernation on the tracking performance, we compared the mean line lifetime with and without hibernation on the real powerline dataset.
The tracker without hibernation corresponds to our implementation of~\cite{everding2018low}, which ignores event polarity and includes the initialization block.
With hibernation, the mean lifetime of a line is 31.12 s.
Without hibernation, the lines exists on average for 2.97 s. The detected lines without hibernation are depicted in Fig.~\ref{fig:power_line_no_hib}. 
The higher line IDs in~\cite{everding2018low} show that the proposed hibernation improves the persistence of the tracker. 
As it is usual during powerline inspection, the dataset contains a lot of slow direction changes.
Thus, there are some prolonged periods where just few events are emitted. 
\textit{Hibernated} lines help dealing with those eventless periods.

\begin{figure}[tb]
    \centering
    \includegraphics[width=0.324\linewidth]{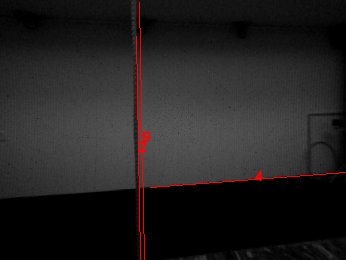} \hfill
    \includegraphics[width=0.324\linewidth]{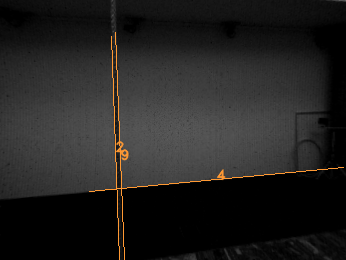} \hfill
    \includegraphics[width=0.324\linewidth]{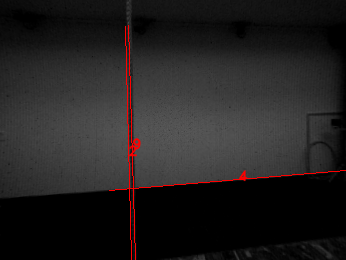}
    \\[\smallskipamount]
    \includegraphics[width=0.324\linewidth]{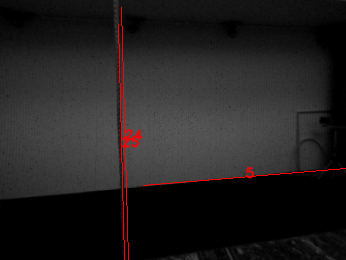} \hfill
    \includegraphics[width=0.324\linewidth]{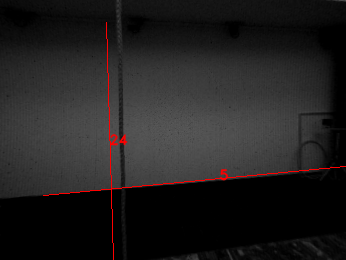} \hfill
    \includegraphics[width=0.324\linewidth]{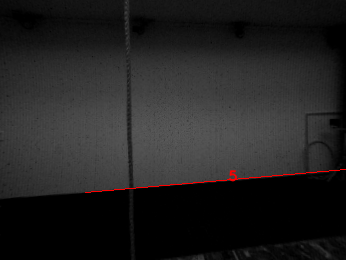}
    
    \caption{Swift change of direction with hibernation (top row) and without hibernation (bottom row). The line moves to the left direction (left column), the line stands still at the exact moment the direction changes (middle column), and the line moves to the right (right column).}
    \label{fig:hibernation_change_dir}
\end{figure}

To illustrate the advantage of hibernation in the case of swift direction changes, the tracker was also evaluated with and without hibernation on a dataset with more aggressive direction changes. 
In this dataset, a rope is suspended from the ceiling whilst the camera yaws from left to right and back in an oscillatory fashion. 
In Fig.~\ref{fig:hibernation_change_dir}, the bottom row shows the issue of swift direction changes. 
With hibernation, in the top row, the line goes into the \textit{hibernated} state at the exact moment when the camera changes its direction and thus will have its midpoint projected orthogonally onto the $xy$ plane. 
Instead, the line without hibernation continues to project its midpoint along the fitted event plane which here keeps moving in the left direction. 
The line is detached from the actual line and is deleted after a short period of time. 

\subsection{Hardware Test}\label{sec:hardware_test}

\begin{figure}[tb]
    \centering
    \includegraphics[width=0.324\linewidth]{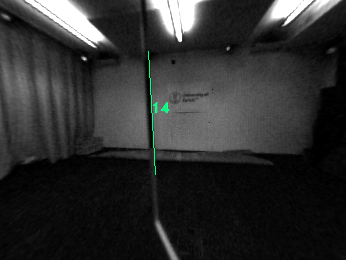} \hfill
    \includegraphics[width=0.324\linewidth]{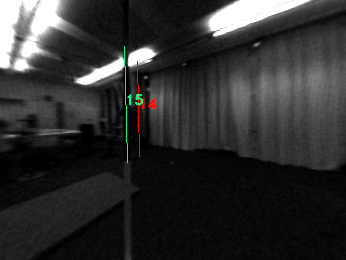} \hfill
    \includegraphics[width=0.324\linewidth]{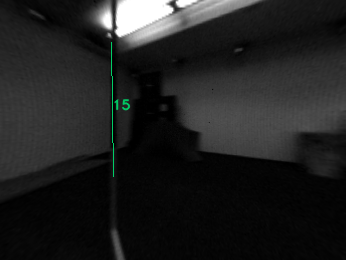} 
    \\[\smallskipamount]
    \includegraphics[width=0.324\linewidth]{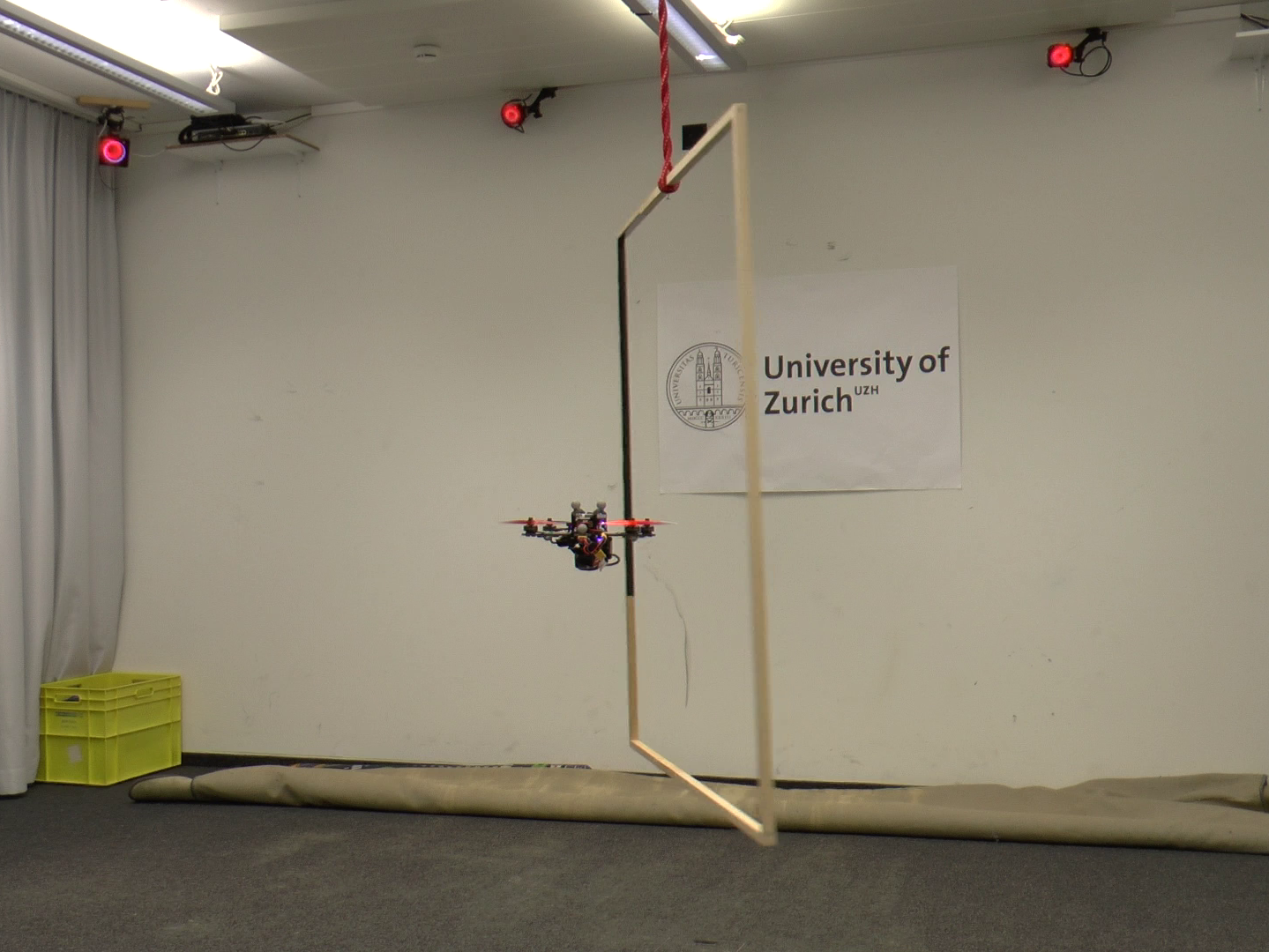} \hfill
    \includegraphics[width=0.324\linewidth]{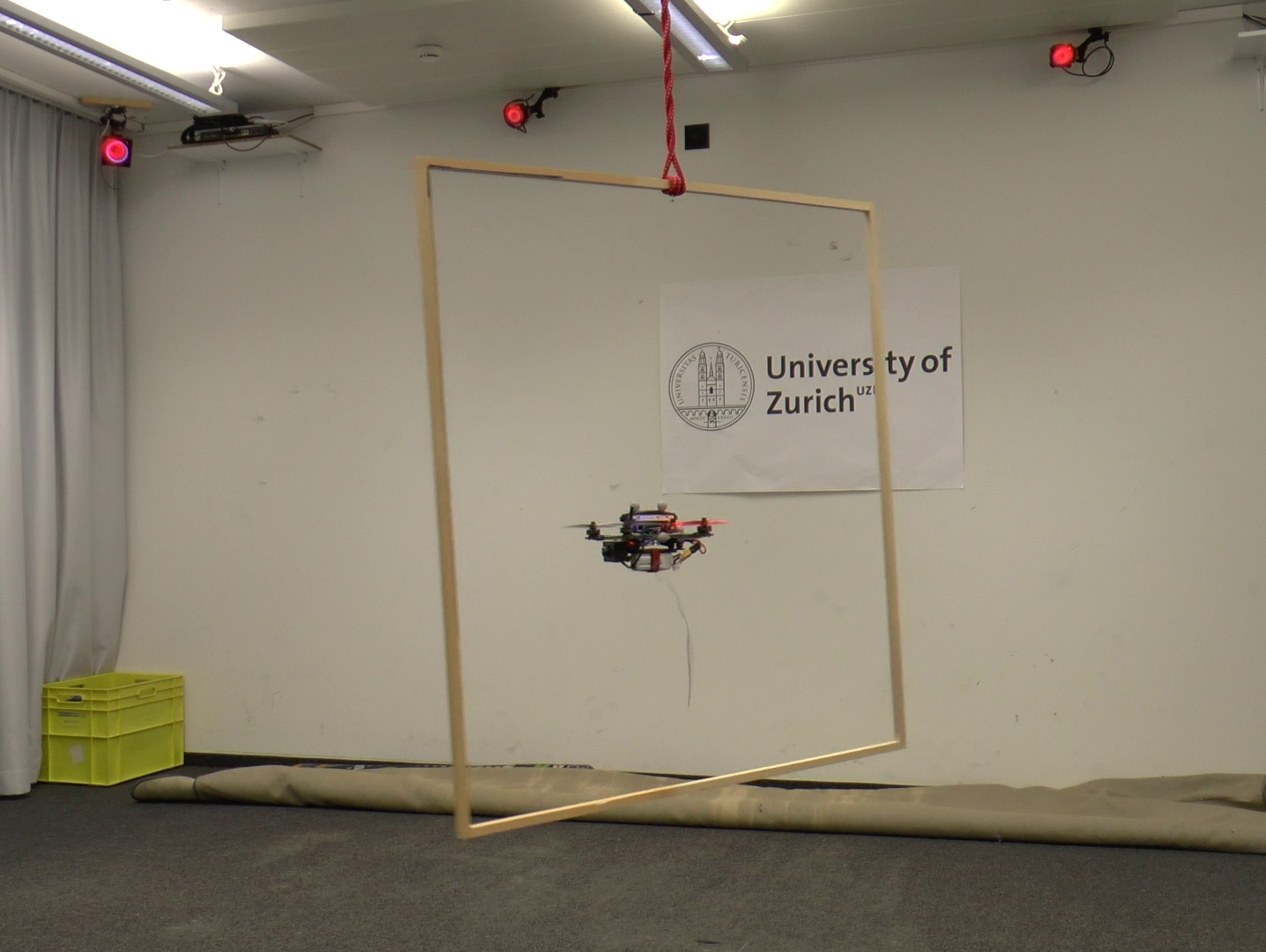} \hfill
    \includegraphics[width=0.324\linewidth]{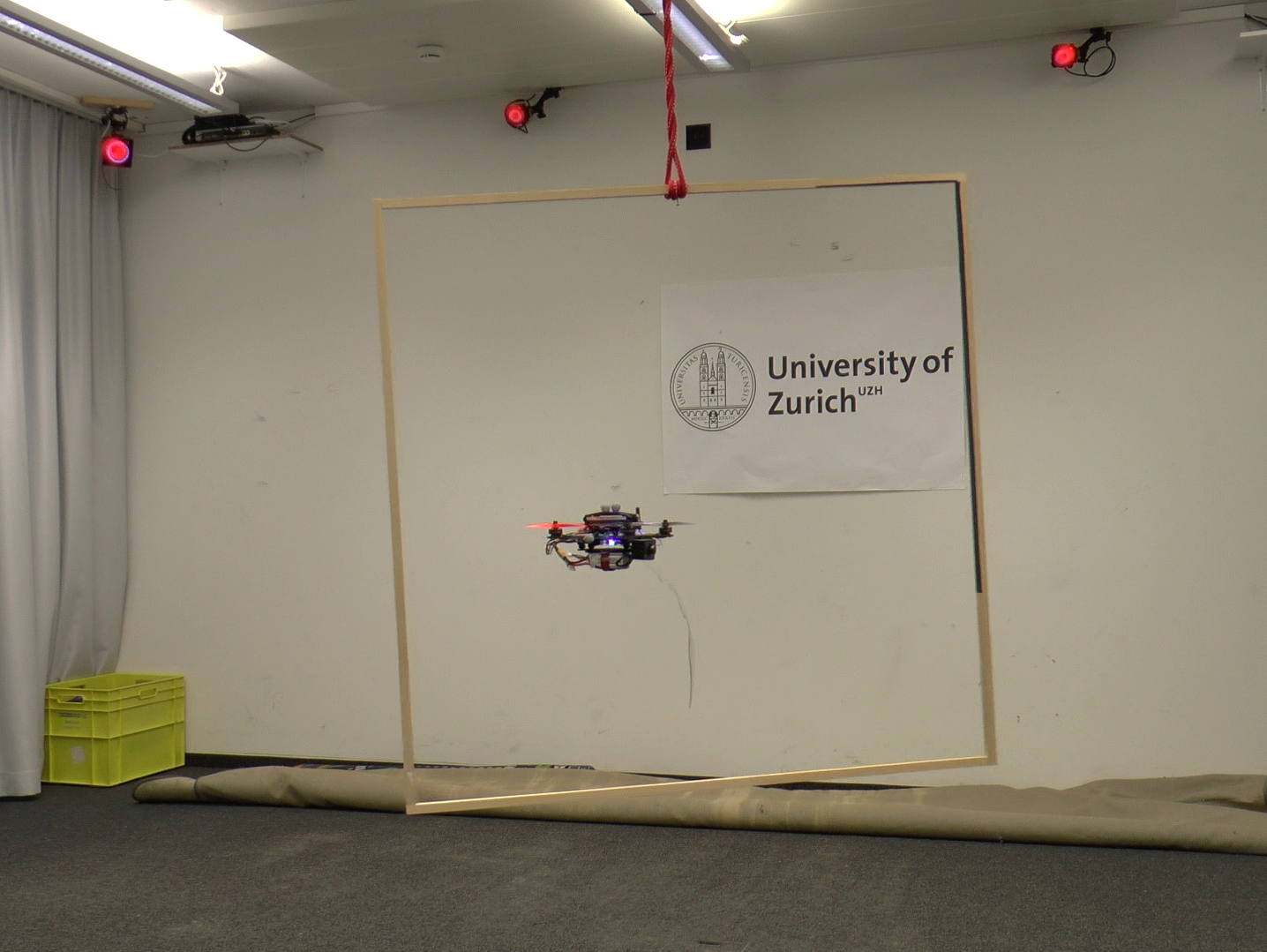} 
    
    \caption{Rotating square followed by the quadrotor. The drone yaws counterclockwise from left to right.}
    \label{fig:rot_square}
\end{figure}

In this set of experiments, we show that our tracker can run onboard a resource quadrotor platform in a closed-loop controller and analyze the computational cost of the different parts of the proposed algorithm.
The platform used for these experiments was a quadrotor equipped with a Nvidia Jetson TX2.
To record the events, the DAVIS 346 camera was used. 
The quadrotor setup is depicted in Fig.~\ref{fig:eyecatcher} (top picture). 
The state estimation was provided by an external motion capture system. 
The line tracker ran onboard together with a model predictive controller (MPC)~\cite{falanga2018pampc}. 
The desired yaw rate was computed by a PD controller. The error fed into the controller was the pixel difference between the center of the image and the x coordinate of the midpoint of the reference line.
In this way, we wanted to keep the reference line in the center of the image.
The reference line was selected as the longest vertical line in the scene.
The drone was placed in the middle of a wooden square suspended from the ceiling by a rope. 
The square was then wound up and released. 
The quadrotor tried to follow one of the sides of the rotating square. 
In Fig.~\ref{fig:rot_square}, the top row shows the captured images with the detected lines overlaid. 
The green line is the reference line that the drone tried to follow. 
The drone successfully followed the square for multiple rotations as shown in the accompanying video.

\subsubsection{Computational Cost}\label{sec:computational_cost}

This section evaluates the computational cost of the tracker running onboard the quadrotor platform. 
We aimed to get an estimate of the bandwidth of the incoming events that the tracker can handle. 
The mean processing times of the different processing steps of the proposed line tracker are in Table~\ref{tab:processing_time}.
\begin{table}[tb]
\centering
\begin{tabular}{|l|r|}
\hline
\textbf{Processing step}  & \multicolumn{1}{l|}{\textbf{Processing time {[}ns{]}}} \\ \hline
Filtering        & 276.6                                         \\ \hline
Line addition    & 1,369.8                                        \\ \hline
Cluster addition & 744.0                                         \\ \hline
Cluster creation & 585.7                                         \\ \hline
\end{tabular}

\caption{Mean processing time per event for each step of the proposed line tracker algorithm.}
\label{tab:processing_time}
\vspace{-0.5cm}
\end{table}

\begin{figure}[bt]
    \centering
    \includegraphics[width=1.0\linewidth] {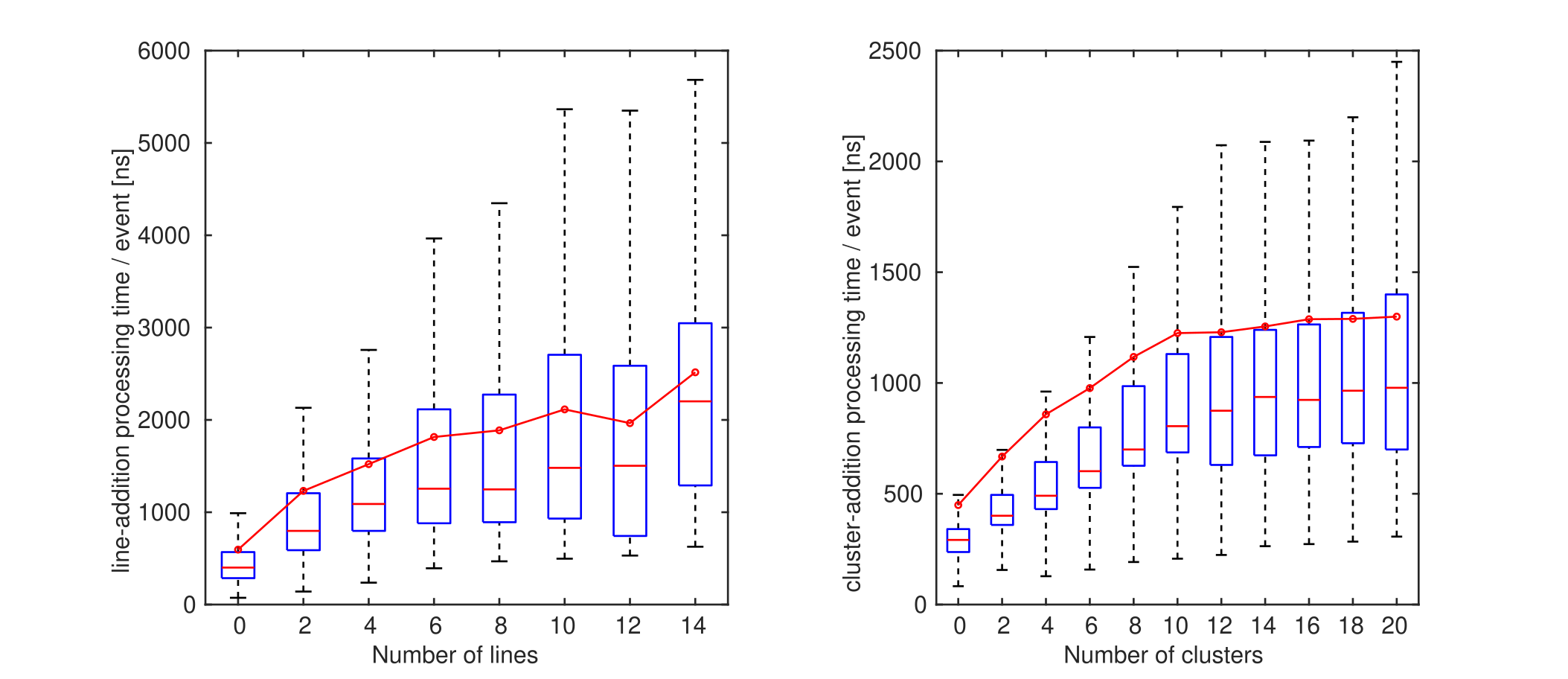}
    
    \caption{Processing time of the line addition and cluster addition steps as a function of the number of lines and clusters, respectively. The red line connects the means.}
    \label{fig:process_time_add_line_cluster}
\end{figure}
Most time is spent trying to add an event to an existing line, followed by the time needed to add an event to a cluster. 
The time needed for the filtering step and the cluster-creation step is independent of the number of lines and clusters. 
The line addition step and the cluster addition step are linearly dependent on the number of lines and clusters, respectively, as shown in Fig.~\ref{fig:process_time_add_line_cluster}. 
The large variance of the data can be explained by the mutex locks. The lines and clusters are objects that are used in the periodical update, which runs on a different thread. 
Whenever a periodical check occurs, the line addition step and the cluster addition step are blocked.
To calculate the bandwidth that the tracker can handle, an assumption of the expected number of lines and clusters has to be made. 
Since our tracker is primarily designed for powerline inspection tasks, the dataset of the real powerline was used to estimate the expected number of lines and clusters, which is 3 lines and 4 clusters. 
The mean processing time for the line addition step is 1.376 $\mu$s, and for the cluster addition step is 0.859 $\mu$s. 
Thus, the overall mean processing time for environments similar to the real powerline data set is 3.097 $\mu$s. 
This means the tracker can handle approximately 320,000 events per second. 
During the hardware tests in Section~\ref{sec:hardware_test} the incoming events stream briefly exceeded 400,000 events per second with 4 lines and 8 clusters. 
Thus, for a short period of time, the tracker was not able to keep up with all the incoming events. 
\section{Conclusion}
\label{sec:conclusion}
This paper presented a powerline tracker for quadrotors using event cameras. 
We proposed to add a new line state, hibernation, to cope with direction changes of the lines.
We showed that hibernation increases the tracker persistence, up to $10\times$, in terms of the lifetime of the detected lines with respect to the state of the art. 
In addition, we showed that our algorithm is able to run onboard resource constrained platforms as quadrotors, and tested it in a closed-loop controller experiments where the quadrotor needs to keep a reference line in the center of the image.
As future work, we plan to include the proposed line tracker in a perception-aware planner.

\section*{Acknowledgment}
\label{sec:acknowledgment}

We would like to thank Christian Pfeiffer and Juan Simon Jaime for their help with piloting the quadrotor to record the powerline dataset.
We would also like to thank Philipp Foehn for his help with the MPC controller.

{\small
\bibliographystyle{IEEEtran}
\bibliography{all}
}

\section{Appendix}
\label{sec:appendix}

\subsection{Main tracker parameters}

\begin{table}[H]
\begin{tabular}{|l|l|l|}
\hline
\textbf{Cluster parameter}      & \textbf{Value} & \textbf{Description} \\\hline
creation number events  & 7 & \begin{tabular}[c]{@{}l@{}} min number of events to create\\ a cluster from a chain.\end{tabular} \\ \hline
addition threshold      & 1.3 pixel           & \begin{tabular}[c]{@{}l@{}} max distance to inferred line for\\ an event to be added\end{tabular}                            \\ \hline
merge angle             & 15 deg             & \begin{tabular}[c]{@{}l@{}} max angle difference between\\ clusters to be merged\end{tabular}                                \\ \hline
cleanup event age       & 50 ms           & \begin{tabular}[c]{@{}l@{}} max age of events before they\\ are removed\end{tabular}                                         \\ \hline
deletion no events      & 40 ms             & \begin{tabular}[c]{@{}l@{}} if there have not been any event\\ for longer than this time, the\\ cluster is deleted\end{tabular}  \\ \hline
\end{tabular}
\end{table}
\begin{table}[H]
\begin{tabular}{|l|l|l|}
\hline
\textbf{Line Parameter}      & \textbf{Value} & \textbf{Description}    \\ \hline
promotion threshold     & 1.2 pixel           & \begin{tabular}[c]{@{}l@{}} value of the\\ smallest eigenvalue\\ for line promotion\end{tabular}                                  \\ \hline
promotion number of events & 35 & \begin{tabular}[c]{@{}l@{}} minimum number\\ of events in cluster\\ for line promotion\end{tabular}                              \\ \hline
initialization length   & 70 pixel            & \begin{tabular}[c]{@{}l@{}} minimum length of\\ the line to pass\\ initialization\end{tabular}                                     \\ \hline
initialization period   & 90 ms             & \begin{tabular}[c]{@{}l@{}} initialization period\end{tabular}                                                                \\ \hline
addition threshold      & 1.8 pixel          & \begin{tabular}[c]{@{}l@{}} max distance to the\\ inferred line for an\\ event to be added\end{tabular}                               \\ \hline
merge angle             & 8 deg             & \begin{tabular}[c]{@{}l@{}} max angle\\ difference between\\ lines to be merged\end{tabular}                                  \\ \hline
merge distance          & 3.5 pixel           & \begin{tabular}[c]{@{}l@{}} max distance of the\\ line mid point to\\ the other line\\ to be merged\end{tabular}                              \\ \hline
hibernation density     & 0.08 $\frac{\text{\# events}}{\text{pixel}^2 \cdot \text{ms}}$          & \begin{tabular}[c]{@{}l@{}} min line density\\ before being\\ hibernated\end{tabular} \\ \hline
cleanup event age       & 50 ms             & \begin{tabular}[c]{@{}l@{}} max age of the events\\ before they are\\ removed\end{tabular}                                         \\ \hline
deletion no events      & 200 ms           & \begin{tabular}[c]{@{}l@{}} if there have not been\\ any events \\ for longer than this, \\ the line is deleted\end{tabular}    \\ \hline
\end{tabular}
\end{table}

\end{document}